\documentclass[a4paper,11pt]{article}
\usepackage{latexsym}
\usepackage{amsmath}
\usepackage{amssymb}
\usepackage{verbatim}
\usepackage{natbib}
\usepackage[colorlinks,citecolor=blue,urlcolor=blue,filecolor=blue,backref=page]{hyperref}

\newif\ifpdf
\ifx\pdfoutput\undefined
  \pdffalse                     
  \usepackage[dvips]{graphicx}
\else
  \pdfoutput=1                  
  \pdftrue
  \usepackage[pdftex]{graphicx}
\fi

\newcommand{\be}{\begin{equation}}
\newcommand{\ee}{\end{equation}}
\newcommand{\bi}{\begin{itemize}}
\newcommand{\ei}{\end{itemize}}
\newcommand{\bea}{\begin{eqnarray}}
\newcommand{\eea}{\end{eqnarray}}

\newcommand{\bfalpha}{\boldsymbol{\alpha}}

\newcommand{\bfpi}{\boldsymbol{\pi}}

\newcommand{\bftheta}{\boldsymbol{\theta}}

\newcommand{\bfe}{\mathbf{e}}

\newcommand{\bfx}{\mathbf{x}}

\newcommand{\tbx}{\tilde{\bfx}}

\newcommand{\cut}[1]{}

\newcommand{\defeq}{{\stackrel{\mathrm{def}}{=}}}

\newcommand{\tQ}{\tilde{Q}}
\newcommand{\tpi}{\tilde{\pi}}
\newcommand{\ttheta}{\tilde{\theta}}
\newcommand{\nb}{Na\"{\i}ve Bayes}

\title{
Na\"{\i}ve Bayes Classifiers and One-hot Encoding of Categorical Variables}

\author{Christopher K. I. Williams, \\
School of Informatics, University of Edinburgh \\
{\tt c.k.i.williams@ed.ac.uk}}

\date{\today}

\begin{document}
\maketitle

\begin{abstract}
  This paper investigates the consequences of encoding a $K$-valued
  categorical variable incorrectly as $K$ bits via one-hot encoding,
  when using a \nb\ classifier. This gives rise to a
  product-of-Bernoullis (PoB) assumption, rather than the correct
  categorical \nb\ classifier. The differences between the two
  classifiers are analysed mathematically and experimentally.
  In our experiments using probability vectors drawn from a Dirichlet
  distribution, the two classifiers are found to agree on the
  \emph{maximum a posteriori} class label for most cases,
  although the posterior probabilities are usually greater for the
  PoB case.
\end{abstract}
    
Consider a \nb\ classifier that uses a categorical variable $x$ in
order to predict a class label $y$. An example of such a variable $x$ is
eye colour, which may take on values brown, gray, blue, green
etc. Suppose that $x$ can take on $K$ different values, coded as $1,
\ldots, K$.  The correct way to make use of this variable in a
\nb\ classifier is to estimate $p(x=j|y)$ for $j = 1, \ldots, K$ for
each possible value of $y$; see, e.g., \citet[sec.\ 9.3]{murphy-22}.
However, sometimes the categorical variable may be recoded as $K$ bits
in a \emph{one-hot} encoding scheme (see, e.g.,
\citealt*[sec.\ 1.5.3.1]{murphy-22}), also known as ``dummy
variables'' in the statistics literature. For example, this is a
common input encoding for use with neural networks. If these bits are
na\"{\i}vely treated as $K$ independent Bernoulli variables, then the
classification probabilities will not be correctly computed under the
\nb\ model. (Clearly the $K$ bits are not independent, as if one is
on, all the rest are off).\footnote{ Note also that there are $2^K$
possible states of $K$ bits, but only $K$ of these are one-hot codes.}

In {\tt scikit-learn} \citep{scikit-learn-11}, the 
{\tt sklearn.naive\_bayes} classifier will work correctly if the {\tt
  OrdinalEncoder} is used to encode categorical variables, but not if
the {\tt OneHotEncoder} is used for them. And, for example, the {\tt
  fastNaiveBayes} package \citep[p.\ 4]{skogholt-20}
``will convert non numeric columns to one hot encoded
features to use with the Bernoulli event model'', i.e.\
not the correct behaviour for categorical variables. But
it does recognize categorical variables if they are coded as integers.

Let $p(x=j|y=i)$ be denoted by $\theta_{ji}$, and the one-hot
encoding of $x$ be denoted by $\tilde{\bfx}$. Let
$\tilde{\theta}_{ji}$  denote $p(\tilde{x}(j)=1|y=i)$, 
the marginal probability of the $j$th bit in $\tilde{\bfx}$ being 1.
Note that there are $C(K-1)$ free parameters in $\Theta$ due to
normalization, but $CK$ in $\tilde{\Theta}$.
Let $\bfe_j$ denote the one-hot vector with a 1 in the $j$th 
position. Thus assuming that the \nb\ assumption applies
across these bits, we have
\begin{equation}
p(\tilde{\bfx}= \bfe_j|y=i) = \tilde{\theta}_{ji} \prod_{k \neq j} (1 -
\tilde{\theta}_{ki}) \defeq  \tQ^{-j}_i \tilde{\theta}_{ji} . \label{eq:full}
\end{equation}

Under the correct categorical model, and assuming there is only
one feature $x$, we have that
\begin{equation}
 p(y=i|x=j)  = \frac{ \pi_i \theta_{ji}}{\sum_{c=1}^C \pi_c
   \theta_{jc}} , \label{eq:post-cat}
\end{equation}
where $\bfpi = (\pi_1, \ldots, \pi_C)$ is the vector of prior
probabilities over the classes.

Under the product-of-Bernoullis (PoB) assumption, we have that
\begin{equation}
 p(y=i|\tilde{\bfx}=\bfe_j)  = \frac{ \tpi_i \tQ^{-j}_i \ttheta_{ji}}
{\sum_{c=1}^C \tpi_c \tQ^{-j}_c \ttheta_{jc}} , \label{eq:post-pob}
\end{equation}
where $\tilde{\bfpi} = (\tpi_1, \ldots, \tpi_C)$ is the vector of prior class
probabilities for this model. If there was more than one categorical
input feature, there will be a $\tQ^{-j}$ factor coming
from each one. But factors corresponding to 
true Bernoulli features, or to Gaussian models for real-valued
features in a \nb\ classifier  would be unaffected by one-hot
encoding.

If the parameters of the product-of-Bernoullis model are estimated
by maximum likelihood, we will have that $\tilde{\bfpi} = \bfpi$, and
that $\tilde{\Theta} = \Theta$. 
This is because the prior class probabilities are the same for the PoB
model as for the categorical model, and the probability of the $j$th
bit in $\tbx$ being set to 1 given $y=i$ is $\theta_{ji}$. Given this,
we drop the tildes on $\bfpi$, $\Theta$ and $Q$ factors below.
The differences between eqs.\
\ref{eq:post-cat} and \ref{eq:post-pob} are now that there is an
additional factor of $\tilde{Q}^{-j}_i$ (resp.\ $\tilde{Q}^{-j}_c$) 
in the numerator (resp.\ denominator) of eq.\ \ref{eq:post-pob}.

Below we first analyze $Q^{-j}$ as a function of its parameter vector,
and then consider the consequences of the change from
eq.\ \ref{eq:post-cat} to eq.\ \ref{eq:post-pob} on the
\nb\ classifier. Experiments to explore this further are shown
in sec.\ \ref{sec:expts}. We conclude with a discussion.

\section{Analysis of $Q^{-j}$ \label{sec:Q-j}}
In this section we suppress the dependence of $Q$ on the class label,
and consider a parameter vector $\bftheta = (\theta_1, \ldots, \theta_K)$.
Thus we consider 
\begin{equation}
Q^{-j}(\bftheta) = \prod_{k \neq j} (1 - \theta_k)  ,
\end{equation}
subject to the constraint that $\sum_{k=1}^K \theta_k = 1$ and that
the $\theta$s are non-negative. We explore how $Q^{-j}(\bftheta)$
varies as a function of $\bftheta^{-j} = (\theta_1, \ldots,
\theta_{j-1}, \theta_{j+1}, \ldots, \theta_{K})$ for
a fixed value of $\theta_j$. $\bftheta^{-j}$ lives on the simplex
defined by the constraint $\sum_{k \neq j} \theta_k = 1 - \theta_j$.

\begin{figure}
\begin{tabular}{cc}
   \includegraphics[width=.5\textwidth]{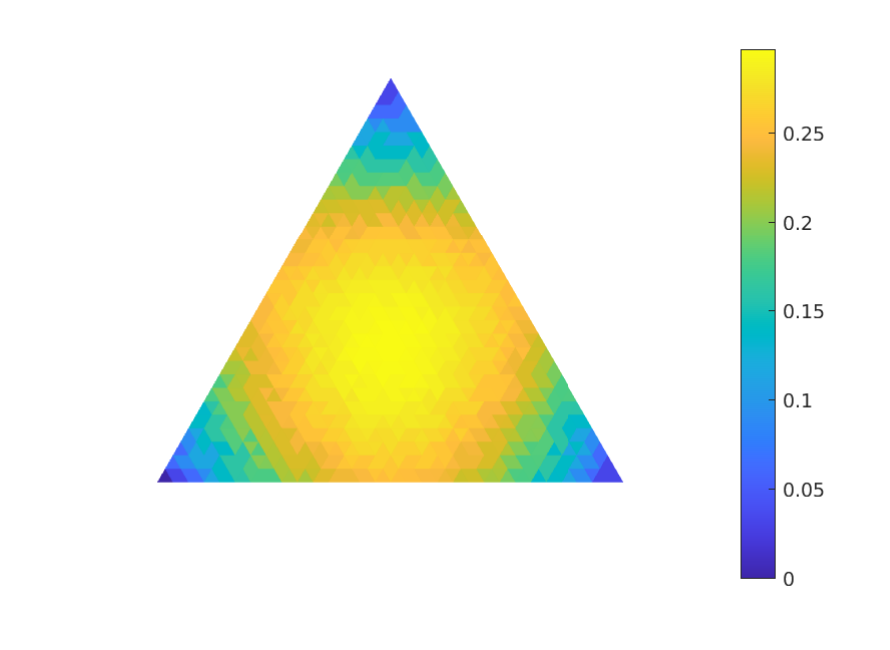} &
   \includegraphics[width=.5\textwidth]{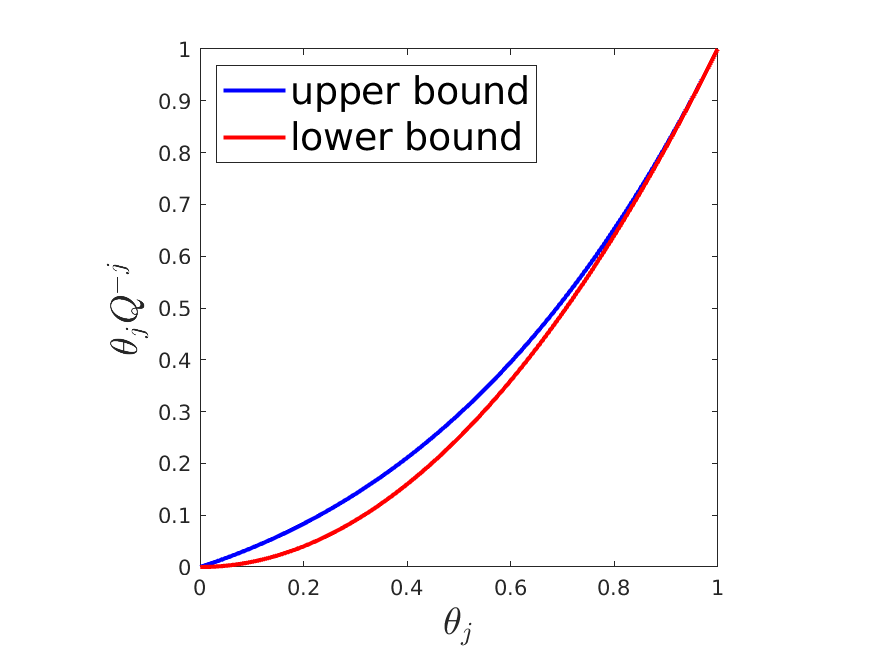} \\
  (a) & (b) 
\end{tabular}
\caption{(a) Plot of $Q^{-4}(\theta_1,\theta_2,\theta_3)$
on the simplex, for $\theta_4 =0$.
(b) Upper and lower bounds on $f_j(\bftheta)$ against $\theta_j$
  for $K=6$. \label{fig:Qul}}
\end{figure}

$Q^{-j}(\bftheta)$ is a symmetric function of the elements of
$\bftheta^{-j}$. A plot of the 4-dimensional case with $\theta_j=0$ is
shown in Fig.\ \ref{fig:Qul}(a). Note that the minimum values occur
in the corners of the simplex (where only one of the $\theta$s is
non-zero), and the maximum in the centre, where they all the variables
are equal.

For the general $K$-dimensional case it is thus natural to investigate the
value at the corners and centre of the simplex. At a corner we have
$\theta_k = 1 - \theta_j$ for a single vertex $k$, with the remaining
$K-2$ $\theta$ values being 0. In this case we have $Q^{-j}(\bftheta)
= 1 - (1-\theta_j) = \theta_j$.

We now use calculus to find the optimum/optima of $Q^{-j}(\bftheta)$ subject
to the constraint $\sum_{k \neq j} \theta_k = 1 - \theta_j$. Defining
the Lagrangian $J = Q^{-j}(\bftheta) + \lambda (\sum_{k=1}^K \theta_k
- 1)$, we have that 
\begin{equation}
\frac{\partial J}{\partial \theta_i} = - \prod_{k \neq i, \; j}  (1 -
\theta_k) + \lambda \qquad  \mathrm{for}\;
i = 1, \ldots, j-1, j+1, \ldots, K.
\end{equation}
Setting this equal to 0 and multiplying through by $(1-\theta_i)$,
we obtain 
\begin{equation}
(1 - \theta^*_i) = \frac{Q^{-j}(\bftheta^*)}{\lambda} \qquad
  \mathrm{for} \; i = 1, \ldots, j-1, j+1, \ldots, K,
\end{equation}
where $\theta^*_i$ denotes the value of $\theta_i$ at the optimum
(and similarly for the other variables in
$\bftheta^{-j}$). $\bftheta^*$ denotes the vector $(\theta_1^*,
\ldots, \theta^*_{j-1}, \theta_j, \theta^*_{j+1}, \ldots, 
\theta_{K}^*)$.
Hence the optimum lies at the centre of the simplex, where
$\theta^*_i = (1 - \theta_j)/(K-1)$ for $i \neq j$.
Hence
\begin{equation}
1 -  \theta^*_i = 1 - \frac{1 - \theta_j}{K-1} = \frac{K-2 + \theta_j}{K-1},
\end{equation}
and at the optimum 
\begin{equation}
  Q^{-j}(\bftheta^*) = \prod_{k \neq j} (1 - \theta^*_k) = \left[
    \frac{K-2 + \theta_j}{K-1} \right]^{K-1} . \label{eq:Qtheta*}
\end{equation}  
 Analysis of $Q^{-j}(\bftheta^*)$ as a function of $\theta_j$ shows
 that it is greater than $\theta_j$ for $0 \le \theta_j < 1$, with
 equality at $\theta_j =1$ (when all the other $\theta$s must all be
 zero).
 To see this, at $\theta_j = 0$ we have that $Q^{-j}(\theta_1^*,
\ldots, \theta^*_{j-1}, 0, \theta^*_{j+1}, \ldots, \theta_{K}^*) = \left[
   \frac{K-2}{K-1} \right]^{K-1} > 0$, and the derivative of
 $Q^{-j}(\bftheta^*) > 0$ in the interval,  so it is monotonically increasing.
 Hence we  conclude that {\bf $Q^{-j}(\bftheta)$ is bounded between its
 minimum value
 of $\theta_j$} at the corners of the simplex, {\bf and its maximum value of 
 $Q^{-j}(\bftheta^*)$} (as per eq.\ \ref{eq:Qtheta*}) obtained at the centre.
 
\section{The effect of the $Q^{-j}$ factors on the \nb\ classifier}
The difference between the the posterior probabilities computed under
the categorical model of eq.\ \ref{eq:post-cat} and the PoB of
eq.\ \ref{eq:post-pob} is the $Q^{-j}$ factors that appear in the
numerator and denominator of the latter.  The above analysis shows that
$f_j(\bftheta) \defeq \theta_j Q^{-j}(\bftheta)$
can be bounded from above and below.
Let these upper and lower bounds \cut{on $f_j(\bftheta)$} be denoted by
$u(\theta_j)$ and $\ell(\theta_j)$ respectively, with
\begin{align}
  \ell(\theta_j) &= \theta_j^2, \\
  u(\theta_j)    &= \theta_j \left[\frac{K-2 + \theta_{j}}{K-1} \right]^{K-1}.
\end{align}
Fig.\ \ref{fig:Qul}(b) illustrates these bounds for $K=6$.  It is
notable that the gap between the upper and lower bounds is small, and
it is even smaller for lower $K$.

Notice that both bounds pass through the origin and are
convex functions.  Elementary analysis shows that, for $b > a$
\begin{equation}
  \frac{\ell(b)}{\ell(a)} >   \frac{b}{a} \qquad \mathrm{and} \qquad
  \frac{u(b)}{u(a)} >   \frac{b}{a} 
  \label{eq:ratio}
\end{equation}
for $0 < a, \; b < 1$.

For two classes $c$ and $d$, if $\theta_{jc} > \theta_{jd}$, is it true
that $f_j(\bftheta_c) > f_j(\bftheta_d)$? In general this does not
hold, as if $\theta_{jc}$ and $\theta_{jd}$ are sufficiently close,
then it could be that $u(\theta_{jd}) > \ell(\theta_{jc})$. However,
if $\theta_{jc}$ is sufficiently large relative to $\theta_{jd}$, then
we will have that $\ell(\theta_{jc}) > u(\theta_{jd})$ and thus
$f_j(\bftheta_c) > f_j(\bftheta_d)$ will hold.

Applying eq.\ \ref{eq:ratio} to the two classes with $\theta_{jc}$ and
$\theta_{jd}$, we have that for both the upper and lower bounds, the
ratio of the bounds is greater than $\theta_{jc}/\theta_{jd}$. This
suggests that the introduction of the $Q^{-j}$ factors might make the
ratio $f_j(\bftheta_c) / f_j(\bftheta_d)$ more extreme than
$\theta_{jc}/\theta_{jd}$. This is backed up, on average, in the
experiments in section \ref{sec:expts}.

We can make use of the upper and lower bounds to determine when the
ratio $f_j(\bftheta_c) / f_j(\bftheta_d)$ is guaranteed to be more
extreme than $\theta_{jc}/\theta_{jd}$.  We have that
\begin{equation}
\frac{f_j(\bftheta_c)}{f_j(\bftheta_d)} >
\frac{\ell(\theta_{jc})}{u(\theta_{jd})} =
\frac{\theta_{jc}}{\theta_{jd}} \cdot
\frac{\theta_{jc}}{\left[\frac{K-2 + \theta_{jd}}{K-1} \right]^{K-1}
}. \label{eq:fratio}
\end{equation}
So the ratio is guaranteed to be more extreme if the last term in
eq.\ \ref{eq:fratio} is larger than 1, i.e.\ if
\begin{equation}
\theta_{jc} > \left[\frac{K-2 + \theta_{jd}}{K-1} \right]^{K-1} . 
\end{equation}
A similar analysis can also be used to bound
$f_j(\bftheta_c)/f_j(\bftheta_d)$ from below as
$u(\theta_{jc})/\ell(\theta_{jd})$.

The observations above are consistent with the idea that the PoB
assumption ``overcounts'' the evidence from the $x$ variable, relative
to the correct categorical encoding.

\section{Experiments \label{sec:expts}}

In the experiments below we generate \nb\ classifiers with $C$ classes
and $K$ values of $x$. Each $\bftheta$ vector (of length $K$) is generated by
drawing from a Dirichlet distribution $\mathrm{Dir}(\bfalpha)$,
where $\bfalpha = \alpha (1, 1, \ldots, 1)$, and similarly for
each $\bfpi$ vector (of length $C$). 
$\alpha = 1$  gives rise to a uniform distribution across the
probability simplex. $\alpha > 1$ would put more probability
mass in the centre of the simplex, while $0 < \alpha < 1$ gives
rise to a sparse distribution, with ``spikes'' at the corners of the simplex.

\begin{figure}
\begin{tabular}{cc}
  \includegraphics[width=.5\textwidth]{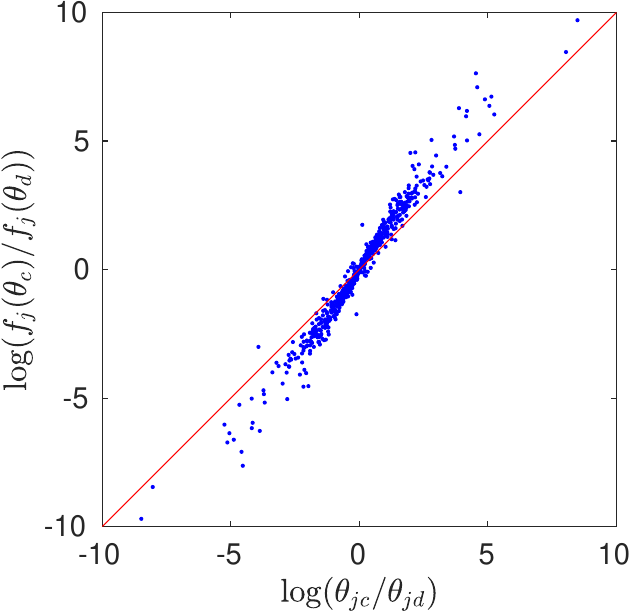} &
  \includegraphics[width=.5\textwidth]{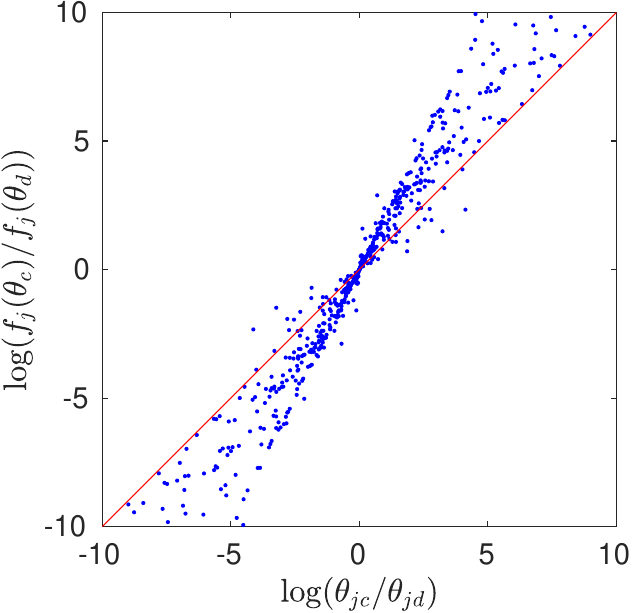} \\
  (a)  $K=3$, $\alpha=1$.  & (b) $K=3$, $\alpha=1/3$. 
\end{tabular}
\caption{(a) A plot of $\log (f_j(\bftheta_c)/f_j(\bftheta_d))$ against
  $\log(\theta_{jc}/\theta_{jd})$ for $K=3$ and $\alpha=1$.  (b)
  The same, but for $K=3$ and $\alpha=1/3$. \label{fig:fratio}}
\end{figure}  

Fig.\ \ref{fig:fratio}(a) shows a scatter plot
of $\log (f_j(\bftheta_c)/f_j(\bftheta_d))$ against
$\log(\theta_{jc}/\theta_{jd})$. This was obtained for $K=3$ and
sampling the $\bftheta$ vectors from the Dirichlet distribution with
parameter $\alpha=1$.\footnote{The plot replicates a point $(x,y)$
at $(-x,-y)$, as the order of the classes $c$ and $d$ is arbitrary.}
Notice that the slope of the scatterplot around
$\log(\theta_{jc}/\theta_{jd})=0$ is greater than 1, agreeing with the
idea that, on average, the introduction of the $Q^{-j}$ factors makes
the ratio $f_j(\bftheta_c) / f_j(\bftheta_d)$ more extreme than
$\theta_{jc}/\theta_{jd}$. There are exceptions to this (i.e.\ points
below the diagonal for a positive log ratio) when the 
log ratio is not too large.

The behaviour of plot of $\log (f_j(\bftheta_c)/f_j(\bftheta_d))$
against $\log(\theta_{jc}/\theta_{jd})$ changes if set $\alpha = 1/3$
(maintaining $K=3$), as shown in Fig.\ \ref{fig:fratio}(b). This value
of $\alpha$ makes the distribution sparser, so that if one value of a
$\bftheta$ vector is large, the others are likely to be small.  If the
ratio $\log(\theta_{jc}/\theta_{jd})$ to be large, $\theta_{jc}$ must
be large and $\theta_{jd}$ small. This is more likely to happen for
$\alpha=1/3$ than for $\alpha=1$, and is reflected in the wider spread
of points in Fig.\ \ref{fig:fratio}(b) compared to panel (a). Another
notable feature in Fig.\ \ref{fig:fratio}(b) is the points lying on the
line running from -5 to 5, corresponding to the function 
$\log((\theta_{jc}/\theta_{jd})^2)$ or
$2 \log(\theta_{jc}/\theta_{jd})$. This line is not a bound, as there are
some points lying above it for positive
$\log(\theta_{jc}/\theta_{jd})$, and below it for a negative log
ratio. To understand this, recall that for the ratio
$\theta_{jc}/\theta_{jd}$ to be large, $\theta_{jc}$
must be large and $\theta_{jd}$ small.  If $\theta_{jc}$ is large,
because of sparsity we would expect one of the other entries in
$\bftheta_c$ to be near zero, and the other to be near
$1-\theta_{jc}$. This would give a $Q^{-j}_c$ factor of around $(1-0) \cdot
(1 - (1 - \theta_{jc}) = \theta_{jc}$. Similarly if $\theta_{jd}$ is
small, we would expect one of the other entries in $\bftheta_d$ to be
near zero, and the other to be near $1-\theta_{jd}$. This would give a
$Q^{-j}_d$ factor of around $\theta_{jd}$. Hence we have that in this
case
\begin{equation}
  \frac{f_j(\bftheta_c)}{f_j(\bftheta_d)} =
\frac{\theta_{jc} \, Q^{-j}(\bftheta_c)}{\theta_{jd} \,
    Q^{-j}(\bftheta_d)} \simeq \left(
    \frac{\theta_{jc}}{\theta_{jd}} \right)^2 ,
\end{equation}
which is in agreement with the points lying on the line of slope 2 in
the plot.

For $K=6$ scatterplots similar to Fig.\ \ref{fig:fratio} for
$\alpha=1$ and $\alpha=1/K$ show a greater tendency for points to
lie closer to the diagonal than for $K=3$. As with Figures
\ref{fig:fratio}(a) and \ref{fig:fratio}(b), the plot for $\alpha=1/K$
exhibits a broader distribution of $\theta_{jc}/\theta_{jd}$ than for
$\alpha=1$, for the same reason. Interestingly there is no obvious
alignment of points on the line of slope 2 for $K=6$ and
$\alpha=1/K$, in contrast to the case for $K=3$.

\paragraph{Consequences of the $f_j(\theta_c)/f_{j}(\theta_d)$ transformation
  for the winning class.} In this paragraph we consider
a two class problem.  For class $c$ to be
the winner under the categorical model when the observed value is $j$,
we have that $\pi_c \theta_{jc} > \pi_d \theta_{jd}$ 
or that $\theta_{jc}/\theta_{jd} > \pi_d/\pi_c$, where $d$ denotes the
other class. Let the ratio
$\pi_d/\pi_c$ be denoted by $\rho$. We now consider two cases when
class $c$ is the winning class  under the categorical model:
\begin{itemize}
\item If $\log \rho$ is somewhat larger than 0, then points on the
  scatterplots lying to the right of the vertical line
$\log(\theta_{jc}/\theta_{jd}) = \log \rho$  will be classified as
  class $c$. Almost all of these points lie above the horizontal line
  $\log(f_j(\bftheta_c)/f_j(\bftheta_d)) = \log \rho$, which means that
  they will also be classified as class $c$, and indeed
  the introduction of the $Q^{-j}$ factors in the PoB model may well make the
posterior in favour of class $c$ more extreme here.
\item If $\log \rho$ is somewhat less than 0, then the prior ratio is strongly in
  favour of class $c$, and one can have $\theta_{jc} < \theta_{jd}$
and  still have class $c$ as the winning class. However, the effect of
the $Q^{-j}$ factors on points lying between $\log \rho$ and 0 on the
x-axis of the plot tends to push them downwards, so the 
points in this region may have a lower posterior probability for class $c$,
or to change classification if they fall below the horizontal line
$\log(f_j(\bftheta_c)/f_j(\bftheta_d)) = \log(\rho)$.
\end{itemize}  

\begin{figure}
\begin{center}  
  \includegraphics[width=.5\textwidth]{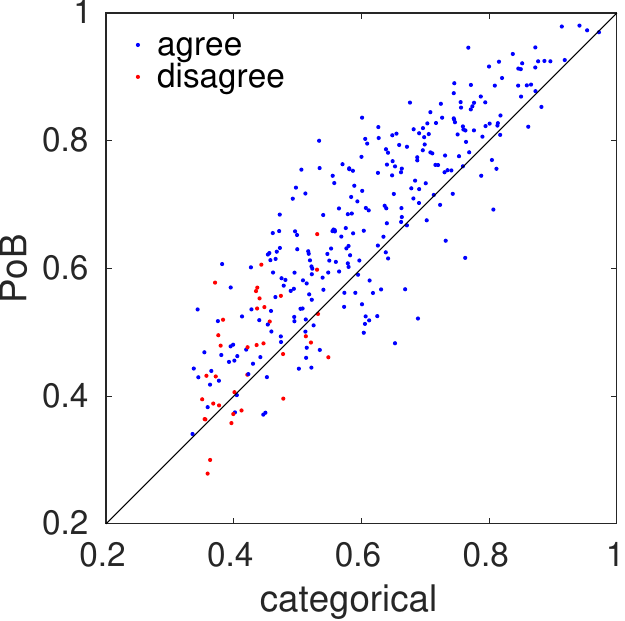}
\end{center}    
\caption{Plot showing the maximum posterior probability for the PoB
  model against the maximum posterior probability for the categorical
  model, for $K=3$ and $\alpha=1$. Datapoints in blue show occasions
  where the MAP class assignment is the same under both models, while
  the red points mark where they disagree. \label{fig:pobvsexact}}
\end{figure}

\paragraph{Comparing posterior probabilities under the categorical
  and PoB models:} \nb\ classifiers with $C=4$ classes were sampled
$n=100$ times, with $K = 3, \; 6$ and $10$ states, for
$\alpha=1$ and $\alpha=1/K$.
For each classifier, each $x = j$ observation was considered, for $j = 1,
\ldots, K$.  Figure \ref{fig:pobvsexact} shows a scatterplot of the
maximum posterior probability under the PoB model plotted against the
maximum posterior probability for the categorical model, for $K=3$ and
$\alpha=1$. 
We see that the maximum probability under the PoB model is usually higher
than under the exact model, in line with the observations above.
This is the case 82.0\% of the samples for $K=3$ and $\alpha=1$. 
Datapoints in blue show occasions where the MAP class assignment is
the same under both models, while the red points mark where they
disagree. Unsurprisingly, differences in MAP classification are
more likely when the MAP value under the categorical model is low.

The amount of scatter around the diagonal line decreases
as $K$ increases. The percentage of cases where the
maximum probability under the PoB model is higher than for the exact
(categorical) model are 82.0\%, 72.3\% and 74.7 \% for
$K = 3,\, 6, 10$ and $\alpha=1$, and 78.0\%, 78.1\% and 76.4\%
respectively for $\alpha = 1/K$.

\paragraph{Comparing the MAP class assignment under the categorical and PoB
  models:} The sampling protocol was used as in the paragraph above.
For $\alpha=1$ and $K=3$ there were 12.33\% of cases where the exact and PoB
classifiers disagreed over the MAP class. This reduced to 5.67\% and
2.50\% for $K=6$ and $K=10$ respectively. For $\alpha=1/K$
the results were 14.33\%, 7.00\% and 6.30\% respectively. 
Thus we observe fewer disagreements between the categorical  and PoB
classifiers as $K$ increases, and that the fraction of  disagreements
is somewhat higher for the sparser $\alpha=1/K$ than for $\alpha=1$.

\section{Discussion}
In this paper we have investigated the differences for
\nb\ classification between using the exact categorical encoding model
(eq. \ref{eq:post-cat}) and the product-of-Bernoullis model
(eq. \ref{eq:post-pob}) arising from one-hot encoding. In our
experiments with one categorical variable, these two classifiers were
found to agree on the MAP class for much of the time, although the
posterior probabilities were usually higher for the PoB model than the
categorical model.

The analysis and experiments above were for one categorical variable
$x$. If there are multiple categorical variables which are one-hot
encoded, then the effects of the transformation from
$\theta_{jc}/\theta_{jd}$ to $f_j(\bftheta_{c})/f_j(\bftheta_{d})$
will multiply up for each variable. If the evidence from each ratio is
all in the same direction then this will magnify the effect, making
the class probabilities more extreme. But if the evidence of the
different ratios conflicts, the effects will tend to cancel out. Also,
note that if there are many features but the evidence provided by each
feature is weak, we will be operating in a region towards the
left-hand side of Fig.\ \ref{fig:pobvsexact}, where errors are more
likely to occur.

Issues with the encoding of variables as identified above illustrate
the importance of a data dictionary or a metadata repository
giving information such as the meaning and type of
each attribute in a table. If given a ``bare'' table we may not
be aware that a categorical variable has been one-hot
encoded,\footnote{It is possible to detect the linear dependence
of one-hot encoded variables by linearly predicting each variable
from the others, as used in variance inflation factor (VIF)
analysis, see, e.g., \citet[sec.\ 11.3.2]{rawlings-pantula-dickey-98}.}
leading to a mis-application of the product-of-Bernoullis model.

\subsection*{Acknowledgements}
I thank Iain Murray for insightful comments about the role of
$Q^{-j}$ in early draft of the manuscript.

\end{document}